\documentclass[10pt, a4paper]{article}

\usepackage[review]{lrec-coling2024} 
\usepackage{makecell}

\title{HarmPot: An Annotation Framework for Evaluating Offline Harm Potential of Social Media Text}

\name{Author1, Author2, Author3} 

\address{Affiliation1, Affiliation2, Affiliation3 \\
         Address1, Address2, Address3 \\
         author1@xxx.yy, author2@zzz.edu, author3@hhh.com\\
         \{author1, author5, author9\}@abc.org\\}

\abstract{
In this paper, we discuss the development of an annotation schema to build datasets for evaluating the offline harm potential of social media texts. We define ``harm potential” as the potential for an online public post to cause real-world physical harm (i.e., violence). Understanding that real-world violence is often spurred by a web of triggers, often combining several online tactics and pre-existing intersectional fissures in the social milieu, to result in targeted physical violence, we do not focus on any single divisive aspect (i.e., caste, gender, religion, or other identities of the victim and perpetrators) nor do we focus on just hate speech or mis/dis-information. Rather, our understanding of the intersectional causes of such triggers focuses our attempt at measuring the harm potential of online content, irrespective of whether it is hateful or not. In this paper, we discuss the development of a framework/annotation schema that allows annotating the data with different aspects of the text including its socio-political grounding and intent of the speaker (as expressed through mood and modality) that together contribute to it being a trigger for offline harm. We also give a comparative analysis and mapping of our framework with some of the existing frameworks.
 \\ \newline  }

\begin{document}

\maketitleabstract

\section{Background and Rationale}

India is a country with a rapidly proliferating social media presence with over 700 million users (including 81\% of teens). 
However, despite massive levels of social media usage, digital media literacy remains low in India. A 2020 survey of a``highly educated online sample” of Indians found that roughly 50\% of the fake news presented to them was judged as ``accurate” or ``very accurate” in their control group \citep{guess2020digital}. The wide reach of social media content, the high prevalence of false or misleading information online, and the extreme communalism/groupthink on social media (for example, see \citet{alzaman2019digital} for a study on communalism in India and \citet{mukherjee2020mobile} for a discussion on the impact of online groupthink on mob violence in India), coupled with low levels of media discernment skills, have exacerbated long-standing social divisions in India \citep{froerer2019religious, banerjee2018caste}.
India has now become a hotbed for online content spurring real-world physical violence. 
Online rumours and hate speech leading to physical violence against targeted communities and the subsequent filming of lynching is no longer uncommon in India \footnote{In 2018, for example, rumours and accusations of certain individuals being child-lifters, primarily
spread on social media led to the five additional instances of mob killings.}. 


Online hate has compounded with pre-existing lines of oppression, incentivizing the publicizing of violence against targeted groups for the sake of gaining public recognition and even praise.
Several incidents of lynching have been triggered by misinformation around caste \citep{scroll_2020, Sajlan_2021}, love-jihad (Muslim men eloping with Hindu women) \cite{NewIndianXpress_2018}, religious desecration etc. There are additional contextual triggers that often cause increased levels of online content and subsequent real-world harm. This includes elections \citep{Deka_2019}, where fake news, rumours, misleading and divisive content are typically spiked for political gains, and global crises like the COVID-19 pandemic \citep{al-zaman2021}, which create a context in which users want “someone to blame,” often unjustly.

In the last few years, over 60 datasets of various sizes and kinds, where a wide variety of abusive language has been annotated, have been released publicly  \citep{Vidgen2020,Poletto2021ResourcesAB}. Existing tools such as Hatebase.org, or the Twitter-backed Hate-Lab or a host of other recent studies have focussed on identifying abusive language \citep{Nobata2016, waseem-etal-2017-understanding}, toxic language \citep{Kolhatkar2020, Kaggle2020}, aggressive language \citep{Haddad2019, kumar-etal-2018-aggression, bhattacharya-etal-2020-developing}, offensive language \citep{Chen2012, mubarak-etal-2017-abusive, Nascimento2019, dePelle2016, schafer-burtenshaw-2019-offence, zampieri-etal-2019-predicting, zampieri-etal-2019-semeval, zampieri-etal-2020-semeval, Kumar2021aggressive, steinberger-etal-2017-cross}, hate speech (several including \cite{Akhtar2019, Albadi2018, Alfina2017, bohra-etal-2018-dataset, Davidson2017, malmasi-zampieri-2017-detecting, schmidt-wiegand-2017-survey, delvignaHate:2017, Fernquist2019, Ishmam2019, sanguinetti-etal-2018-italian}), threatening language \citep{Hammer2017}, or narrower, more specific dimensions such as sexism \citep{Waseem2016, waseem-hovy-2016-hateful}), misogyny, Islamophobia \citep{chung-etal-2019-conan, Vidgen2020Detectingweak}, and homophobia \citep{Akhtar2019}. Some datasets include a combination of these such as hate speech and offensive language \citep{Martins2018, mathur-etal-2018-detecting}), or sexism and aggressive language \citep{bhattacharya-etal-2020-developing}. However, while most of these datasets and frameworks aim to model whether hate or offensive speech has been used or not, there has been no dataset or framework that could directly model the relationship and interdependence of online content and offline incidents of harm and violence. 

In this paper, we discuss the development of a framework - HarmPot - that could be used for annotating the text with textual and contextual information such that the annotated dataset could be used for training models that could predict the offline harm potential of online content. In the following sections, we discuss the detailed annotation schema and annotation guidelines, a comparison with the other popular hate speech schema and finally some details of a new dataset annotated with the data.



\section{The HarmPot Framework}
\label{sec:framework}

``Harm Potential” (HarmPot) could be defined as the potential for an online public post to cause offline, real-world physical harm (i.e., violence). Targeted real-world violence is often spurred by a web of triggers, often combining several online tactics and pre-existing intersectional fissures in the social milieu. 
As such we do not focus on any single divisive aspect (i.e., caste, gender, religion, or other identities of the victim and perpetrators) nor do we focus on just hate speech or mis/dis-information. Rather, 
we focus on marking the harm potential of online content within a specific set of intersectional, contextual factors, irrespective of whether it is hateful or not. The HarmPot framework is designed with the aim of answering the following set of questions with respect to a given text -

\textit{\textbf{Who} is being talked to, \textbf{when}, \textbf{how}, \textbf{why} and all this results in \textbf{what magnitude} of harm potential for the addressee?}

Each of these questions is answered by using a set of parameters, defined in our tagset. We discuss each of these in the following subsections.

\subsection{Magnitude of Harm Potential}

Depending on what kind of offline harm the text could lead to, we define two broad kinds of harm potential -
\begin{itemize}
    \item \textbf{Physical Harm Potential: } It defines the potential of a text to lead to acts of physical violence such as murder, mob lynching, thrashing and beating, etc.
    \item \textbf{Sexual Harm Potential: } It defines the potential of a text to lead to acts of sexual violence such as rape (or rape threats), molestation, sexual harassment, etc.
\end{itemize}

Both of these harm potentials are classified on a scale of 0 - 3, defined below. 

\paragraph{Value 0: } A text will be marked as having `0’ harm potential in the following cases:

\begin{enumerate}
    \item Texts which are a part of the dataset but do not actually relate to any specific incident of violence or larger narrative campaign.
    
    
    \item Texts which are blurbs accompanying links to news reports. 
    
    
    \item Texts that criticise public figures and not protected identities. 
    
\end{enumerate}

\paragraph{Value 1: } A text will be marked as having `1’ harm potential, if it is likely to lead to offline harm in very few, specific contexts but more generally is not expected to trigger incidents of offline harm. The most stereotypical instances of such texts include

\begin{enumerate}
    \item Texts that target communities by using slurs and pejorative terms.
    \item Texts that reinforce negative stereotypes regarding a particular community. 
\end{enumerate}

 \paragraph{Value 2: } A text that is likely to trigger offline harm in most of the contexts - it is only in very specific contexts that it may not be interpreted as a call to violence - is marked as `2' on the harm potential scale. Some of the most stereotypical instances of such texts include 

 \begin{enumerate}
     \item Explicit cases of attack or accusations against communities. 
     \item Justifying violence or discrimination against communities. 
 \end{enumerate}

\paragraph{Value 3: } Any text that has a high potential of triggering offline harm, irrespective of the context that it occurs in is marked as `3' on the harm potential scale. Instances of such texts include 
\begin{enumerate}
    \item Explicit and clear calls to violence against communities or people. 
    \item Explicit and clear attempts to instigate violence against communities or people. 
\end{enumerate}

 The magnitude of harm potential is marked at two levels -
 \begin{enumerate}
     \item \textbf{Text Span: } It is marked in conjunction with specific spans of text that are used to refer to specific identities. It refers to the potential of that specific span of text to trigger offline harm/violence against specific identities (refer to Section \ref{subsec:who} for details).
     \item \textbf{Document: } It is the overall harm potential of the document - generally it is calculated based on the harm potential of individual spans; however, in cases where none of the spans refers to specific identities then an overall harm potential of the document is independently ascertained.
 \end{enumerate}

\subsection{Who is being talked to?}
\label{subsec:who}
This parameter is used to identify the specific types of identities that are `mentioned/referred' (and not necessarily targeted) in a particular `span of text'. We discuss the various ontological types of identities that can be potentially targeted in a text. 
Since this parameter works with the magnitude of harm potential, a text span which simply mentions an identity without targeting it will have `0’ harm potential.

There are three broad annotation instructions that are for this parameter -
\begin{enumerate}
    \item  \textbf{Intersectionality: } If more than one identity of the same individual is referred to (viz. Female Dalit or Pakistani Hindu) then the same span is marked with all the identities and the same harm potential is ascribed in all instances - this is how intersectionality is handled in the framework. If different identities are mentioned in different spans then also different spans will carry the same harm potential, considering it to be an instance of intersectionality.
    \item \textbf{Multiple Identities: } If different identities of different individuals are referred to then they might have different harm potential.
    \item \textbf{Multiple Spans: } If more than one span refers to the same identity of the same or different persons, each span could possibly have different harm potentials.
\end{enumerate}


The framework itself does not enforce a specific set of identities to be marked. However, for the current project, the following non-exhaustive set of identities have been marked in the dataset. If needed, more, less or different kinds of specific subtypes of these categories may also be marked in the text. For each identity and its sub-category, a set of additional guidelines was used for deciding whether its harm potential is `0' or not, as discussed below - if it's not `0' then the guidelines for marking the magnitude of harm potential are to be used.

    \textbf{Caste: } A span is annotated as targeting this identity if there are threats of violence, justification for caste-based discrimination, justification and support for untouchability and criticism of reservation (affirmative action) policy that questions the intellectual capability of these groups. 
    
    \textbf{Religion: } A span is annotated as targeting a member of a religious community if it calls for or justifies violence against them. Propounding or justifying conventional stereotypes associated with the members of such communities or using religious slurs will also have a harm potential greater than `0'. For example, Muslims being called terrorists or jihadis, Muslims and Christians being targeted for alleged forced conversions and Sikhs being called Khalistanis or secessionists. 
    
    \textbf{Descent: } For our specific case, descent encompasses all identities based on inherited status. This includes ethnicity, race, and place of origin (including linguistic or cultural minorities) of a victim (but not caste given its prevalence in the Indian context). 
    Spans supporting or justifying attacks based on places of origin are annotated under this category. 
    
    \textbf{Gender: } This label annotates spans attacking gender minorities (LGBTQIA+ community) and women. Spans propounding or justifying conventional stereotypes or using gendered slurs are also marked with non-zero harm potential.
    
    \textbf{Political Ideology: } Political violence including murder, lynching, thrashing, etc of opposing party members or people of different political ideologies happens regularly. 
    Spans calling for or justifying violence, supporting discrimination or furthering stereotypes against the supporters of a political party or ideology are assigned harm potential greater than `0'. However, a criticism of the political ideologies, political leaders, policies, etc are assigned a `0' harm potential.

\subsection{When is the discourse happening?}

This parameter indicates if a text is posted online in relation to or during a major, public event or happening that might add to its harm potential. 
The harm potential of the content may increase when posted during or before such sensitive occasions and may lead to real-world violence in the form of mob lynchings and even ethnic cleansing. The major categories that we marked under this parameter are discussed below. This parameter is marked at the text/document level and the same text could take multiple labels. Since generally the dates of the events are already well-known, these labels could be mostly assigned automatically and could be seen as a grouping of multiple dates in a single category. 

    \textbf{Riots:} In general, violent public disorders are referred to as riots. In India, in the past few years, hate speeches in social media have made a significant contribution to the amplification of violence during the riots. As such posts related to riots at the time of riots (or otherwise) are likely to have higher harm potential than otherwise.
    
    \textbf{Elections:} Elections in India often see violence by supporters of rival political parties, and they are adopted in various themes such as communalism, terrorism allegations, anti-national, systemized threats and disruption of harmony.  
    
    \textbf{Pandemic}
    
    \textbf{Extremist Attack:} An extremist attack on state forces or the public may also lead to online hate against particular communities. The Pulwama suicide attack of February 2019 in India led to widespread hate speech and real-world violence against Kashmiri Muslims throughout India. 
    
    \textbf{Festivals:} Religious festivals have recently become flashpoints for communal violence with different sides accusing each other of attacking processions or interfering with rituals. Online hate and disinformation often spikes during these situations.
    
    \textbf{Group-Specific State Decisions:} This context pertains to when the government introduces or implements legislation/decisions affecting a particular community. The government’s decisions may be criticised or protested against by the community followed by online and offline attacks by the government's supporters. Recent examples in India have been the Citizenship Amendment Act, Farm Laws and the abrogation of Article 370. 
    
    \textbf{Generic:} These refer to the posts related to the incidents that are recurring in nature (like the previous factors) but generally do not have a fixed or pre-determined start or end time (viz. mob lynching on the suspicion of being child-lifters or those related to cow vigilantism in India).
    
    \textbf{Others:} The posts that do not co-occur with any of the above-mentioned contextual factors - seemingly one-off incidents of hate and violence at no specific time - are marked as others.

\subsection{How is it being said?}

Since we focus on the harm potential of social media content, the methodology developed here is sensitive to the fact that the core objects of study are linguistic events themselves and so it is essential to model the textual features viz its lexical, syntactic and semantic properties that co-occur with the contextual features discussed in the earlier subsections. For the current project, we have defined a set of semantic features (specifically mood and modality) and lexical features (affective expressions) that are marked in the text. Since the other morphosyntactic features could be marked automatically using the earlier existing systems or are implicity learnt by modern transformers-based multilingual models, we have not marked those separately. The labels for this parameter are marked at the span level and generally, but not necessarily, overlap with the spans of the `who' parameter. 

There is an abstract link that can be sketched between the language that a speaker uses to convey harm vis-a-vis how that language is particularly structured to reflect the speaker’s intentions and the speaker’s own evaluation of what they say as possible or necessary. The variation in the speaker's intention and their own evaluation of what they say could have a significant impact on the harm potential of what is being said. These could be modelled using the linguistic categories of mood and modality.
We discuss the different subtypes of these two categories used for annotation below -

    \textbf{Mood Type:} The category of mood is a ``grammatical reflection of the speaker’s purpose in speaking” \citep{kroeger2005analyzing} or an indication of ``what the speaker wants to do with the proposition” in a particular discourse context \citep{bybee1985morphology}. Depending on whether the speaker wants to talk about a situation that has or will actualize in their perspective or whether they want to talk about an event that has not actualized, the grammatical form of the construction changes.  We annotate three broad kinds of mood types -
    \begin{itemize}
        \item \textbf{Realis Mood:} The Realis mood portrays situations as actualized, as having occurred or actually occurring, knowable through direct perception \citep{palmer2001mood}. Indicative mood (that expresses actions that actually did take place, are taking place or will take place) is the canonical bearer of realis mood in a language.
        \item \textbf{Irrealis Mood:} The irrealis portrays situations as purely within the realm of thought, knowable only through imagination \citep{palmer2001mood}. It is used to denote situations or actions that are not known to have happened. Modality-marked constructions, conditionals (that convey dependency of a situation on another situation), counterfactuals (that convey a conditional situation in an alternate reality i.e. a situation that cannot actualise because it is contrary to some fact in the actual world), optatives, hortatives and subjunctives (that express contrary to fact situations) are all grouped under the label of irrealis modality.
        \item \textbf{Neither:} Imperatives (that are used to direct the behaviour of the addressee and get them to act a certain way), interrogatives (that are used to ask some information from the addressee), future-tense marked constructions (that indicate that some event will take place in future as compared to speech time) and negative constructions (that assert that something has not taken place) are marked as `neither’.
    \end{itemize}
    \textbf{Illocutionary Mood:} Illocutionary mood draws upon Austin and Searle’s idea of illocution (what the speaker intends to do via his/her speech) act and encodes speaker intention (in illocution) as a category of mood. There is an expected correlation between a speech act and a sentence type since there is a language-independent tendency for certain illocutionary acts to be mapped onto specific grammatical forms. We have used the following subtypes of illocutionary mood -
    \begin{itemize}
        \item \textbf{Declaratives:} Declaratives can be either `direct' (indicative mood) or `indirect' (mainly using the rhetorical forms). A rhetorical question is an indirect speech act (a mismatch between the sentence type and the intended force) which involves the use of the interrogative form for some purpose other than asking questions \citep{kroeger2005analyzing}. It can be used to indirectly assert something and thereby is an indirect declarative.
        \item \textbf{Interrogatives}
        \item \textbf{Imperatives:} It could be in the form of a command, a request, advice, a plea, permission, an offer or an invitation.
        \item \textbf{Admonitive:} Admonitives are the warnings that a speaker issues to the addressee(s). 
        \item \textbf{Prohibitive:} Prohibitives curtail the addressee’s actions and stop them from engaging with some situation or action.
        \item \textbf{Hortative:} It is used for softened commands or exhortations and so shares properties with imperatives \citep{puskas2018wish}. It is often used with first-person inclusive reference (‘let us…’).
        \item \textbf{Optative:} Sentences in an optative mood express a wish or a desire of the speaker that some situation be brought about.
        \item \textbf{Imprecative:} It indicates that the speaker wishes for an unfavourable proposition to come about.
        \item \textbf{Exclamative}
    \end{itemize}
    \textbf{Modality:} Modality can be viewed as speaker modification of a state of affairs with respect to how the basic event/situation is construed by the speaker. Modalities come in two flavours - whether the speech event is `possible’ or `necessary’ given the particular set of conditions. We have used the following modalities for marking the text spans -
    \begin{itemize}
        \item \textbf{Epistemic:} 
        It deals with ``an estimation, typically but not necessarily by the speaker, of the chances or the likelihood that the state of affairs expressed in the clause applied in the world” \citep{oxfordhandbokmoood}. It is marked for a sentence if that sentence concerns the speaker’s knowing or believing that the state of affairs described in the sentence is possibly (possibility or dubitative) or certainly (necessity) true.
        \item \textbf{Deontic:} Deontic modality can be defined as ``an indication of the degree of moral desirability of the state of affairs expressed in the utterance, typically but not necessarily on behalf of the speaker” \citep{oxfordhandbokmoood}. The conception of morality includes ``societal norms as well as personal ethical criteria of the person responsible for deontic assessment” \citep{oxfordhandbokmoood}. This modality involves an evaluation of the state of affairs that ranges from absolute moral necessity to moral acceptability.
        \item \textbf{Dynamic:} Dynamic modality is concerned with (a) an ability or a capacity ascribed to the participants of the action/ situation; (b) a need/necessity imposed on the participant by external circumstances that lie beyond their control; (c) a possibility/potential or necessity/inevitability inherent in the state of affairs described in the sentence and not related to the participants in that state of affairs.
        \item \textbf{Teleological:} Teleological modality concerns ``what means are possible or necessary for achieving a particular goal” \citep{von2006modality}. 

    \end{itemize}

In addition to these, the presence of affective expressions is marked if there is some word in the text that conveys the speaker’s evaluative attitude or some emotional state towards some part of the information being conveyed by the sentence. 

\subsection{Why is it being said?}
This parameter analyses the discursive role of the text placed within its context and checks for the reason or rationale behind posting the text. It is a direct induction of the `discursive roles' by \citet{kumar-etal-2022-comma} in this framework. We discuss the five categories and their relationship to the magnitude of harm potential here -

    \textbf{Attack:} This label is used when any comment/post poses an attack on any individual or group based on any of their identities. 
    Not all attacks are accompanied by a positive harm potential. For example, criticisms, which are not likely to trigger real-world harm against them are tagged as `attack’ with harm potential `0’. 
    
    \textbf{Defend:} This label is used when any comment/post defends or counter-attacks a previous comment/post. 
    Again not all instances of defend have `0' harm potential - in instances where the defense of the perceived `victim' has the possibility of triggering real-world harm against the attacker, they are marked with non-zero harm potential.
    
    \textbf{Abet:} This label is used when any comment/post lends support and/or encourages an aggressive act which has a harm potential. 
    
    \textbf{Instigate:} This label is used when any comment/post encourages someone to perform an aggressive act. The comment itself may or may not be aggressive but the purpose must be to instigate an act that is potentially harmful in the real world. 
    Instigation happens before the event and its purpose is to trigger or provoke a harmful act unlike abet which occurs during or after the harmful act and its purpose is to praise, support, and/or encourage that act as well as other such acts in the future.
    
    \textbf{Counterspeech:} 
    Texts that diffuse the potentially harmful situation will be tagged as counterspeech. Just as influential speakers can make violence seem acceptable and necessary, they can also favourably influence discourse through counterspeech. 

\section{Data Collection and Annotation}

The framework discussed in Section \ref{sec:framework} is developed over several stages and iterations. In order to test the reliability and validity of the framework, we collected a dataset from different social media platforms and annotated those using the framework. As a first step towards data collection, we focussed on a few incidents of physical harm (riots, lynchings etc.) that had a link to online disinformation and hate campaigns from 2016 -- 2022 \footnote{This time period was decided given the introduction of low-cost data and smartphones by Reliance Jio in 2016 which led to a manifold increase in per-capita data usage}. Finding relevant government-published data related to hate crimes was a challenge as the Indian government stopped collecting data on hate crimes in 2017. Therefore, we decided to use databases from non-governmental organisations like Documentation of the Oppressed (DOTO). This database consisted of a list of over 1,100 incidents of offline hate crimes and violence since 2016. Out of these, we sampled a little over 150 crimes since they had a link to social media discourse. We extracted social media data related to these incidents from different social media platforms viz Twitter, YouTube, Facebook, Telegram and WhatsApp. We also ensured that we collected data both from before and after the incident separately. This approach ensured that we got data that had links to offline harm incidents so they could be considered as potential triggers for the offline harm incident; at the same time, we also got data that might be triggers for a related post-event incident. We collected a total of over 417,000 datapoints in Hindi and English using this methodology \footnote{The complete dataset, along with the hate crimes they were associated with will be released publicly for further research}.

\subsection{Inter-Annotator Agreement}

Approximately 5 - 10 data points were selected from approximately 50 incidents for running the inter-annotator agreement experiments. Each of these was annotated by 3 annotators and Krippendorff's Kappa was calculated for the magnitude of harm potential. The first round of experiments with around 500 data points gave a rather dismal Kappa of 0.25. Following this, we made certain changes to the tagset such as merging different categories to reduce overlap across categories (for example, race, ethnicity and nationality under `Who' were combined into a single category of `Descent') and introducing new categories to better classify different kinds of categories (for example, several new categories under mood and modality were added for a better analysis). We also made changes to the annotation guidelines for clarity. These changes led to significant improvements in the alpha - the second round of experiments gave a final value of 0.53.

While the Kappa value still remained low, for a highly subjective task such as predicting the magnitude of harm potential as reasonably good. We started conducting focus group discussions to understand the reason behind disagreements. As it has been argued earlier as well (for example, \citet{kumar-EtAl:2022:LREC1} and also the Perspectivist Data Manifesto \footnote{\url{http://pdai.info/}}), most of these disagreements seemed reasonable. As such we decided not to push for further agreement - instead, we will be making the disaggregated annotations by different annotators publicly available.

\subsection{Data Annotation}

We have annotated a total dataset of approximately 2,000 data points (taking around 15 - 20 data points from over 100 incidents) to demonstrate the validity of the presented framework. We made use of an online app - LiFE App \citep{singh-etal-2022-towards} - for data annotation since it allowed us to annotate the data simultaneously at the document and the span level. 
Each data point was annotated by 3 annotators working independently.



\section{HarmPot and Other Frameworks: A Comparison}

\begin{table*}
    \centering
    \begin{tabular}{c|ccc}\hline
 \multicolumn{4}{|c|}{\textbf{HASOC Framework}}\\\hline
 & Level A & Level B & Span\\\hline
 HarmPot Harm Potential & \makecell{HOF $\epsilon \{0,1,2,3\}$ \\ NOT $\epsilon \{0,1\}$} & \makecell{Offensive $\epsilon \{0,1,2\}$ \\ Hate $\epsilon \{2,3\}$ \\ Profane $\epsilon \{0,1\}$} & -- \\ \hline 
 HarmPot `Who' & -- & -- &\makecell{$HASOC \subset HarmPot$ \\ HASOC $\epsilon \{1,2,3\}$}\\\hline
 \multicolumn{4}{|c|}{\textbf{OLID Framework}}\\\hline 
 &  Level A &  Level B & Level C\\ \hline 
 HarmPot Harm Potential &  \makecell{OFF $\epsilon \{0,1,2,3\}$ \\ NOT $\epsilon \{0,1\}$} &  \makecell {TIN $\epsilon \{\{1,2,3\} \cup$ \\ \{Caste, Religion...\} \}}& \makecell{\{IND,GRP\} $\epsilon$ \\ \{Caste, Religion...\}}\\ \hline
 \multicolumn{4}{|c|}{\textbf{ComMA Framework}}\\ \hline
 &  Aggression &  Aggression Intensity & Threat and Bias\\ \hline 
 HarmPot Harm Potential & \makecell{OAG $\epsilon \{0,1,2,3\}$ \\ CAG $\epsilon \{0,1,2\}$  \\ NOT $\epsilon \{0,1\}$}  & \makecell{\{PTH, STH\} $\epsilon \{1,2,3\}$ \\ {NtAG, CuAG} $\epsilon \{0,1,2\}$} & -- \\ \hline
 HarmPot `Who' & -- & -- & \makecell{$ComMA \subset HarmPot$ \\ ComMA $\epsilon \{1,2,3\}$} \\\hline
 HarmPot `Why' & \multicolumn{3}{c}{ComMA $\epsilon \{ Attack, Defend, Instigate, Abet, Counterspeech\}$}\\\hline
    \end{tabular}
    \caption{Mapping of HarmPot and Other Frameworks}
    \label{tab:comparison}
\end{table*}

As we mentioned earlier, most of the existing frameworks attempt to only model hateful, aggressive, offensive (or one of the other similar flavours) speech but do not attempt to predict the potential of the text to trigger offline harm incidents. However, such language usage is expected to have some correlation with offline harm. Moreover, prior studies have also pointed out the need to flesh out the interrelationship between different frameworks so as to ensure interoperability and cross-use of datasets annotated with different hate speech frameworks \citep{Poletto2021ResourcesAB,kumar-EtAl:2022:LREC1}. In order to understand this relationship, we carried out a comparative study between our framework and three of the other popular frameworks. We took 500 texts annotated with each of these different frameworks, annotated those with the HarmPot framework and carried out a comparative study. The results of these are discussed in the following subsections.

\subsection{HarmPot and HASOC}
Hate Speech and Offensive Content Identification in Indo-European Languages (HASOC) is a series of workshops/shared tasks that have been held since 2019 and that makes data available for Indo-European languages, marked with hate and offensive labels \citep{Mandl2019,Mandl2020,Mandl2021,hasoc_urdu_maaz_2021,Mandl2022}. 
The first level of the schema distinguishes between Hate and Offensive (HOF) and Not hate and offensive (NOT). The second level of the schema classified HOF into three classes - Hate, Offensive and Profane. In the 2022 and 2023 editions, the second level of the schema was a multiclass annotation indicating Standalone Hate (hate by itself), Contextual Hate (hate in the context of its parent) and Non-hate (not hate by itself).
In the 2023 edition, the task of identifying spans of hate was also introduced in the HateNorm track. We took a total of 500 texts each from the 2019 and 2023 editions of the task and annotated those using the HarmPot framework.
to understand their interrelationship.

The study showed that most of the NOT texts had `0' harm potential but the vice-versa was not necessarily true. On the other hand, probably because of the broad definition of HOF (which includes texts with swear words and profanity), unexpectedly, at least some of the offensive texts carried `1' and even `0' harm potential. At the second level, the mapping becomes clearer as most of the `Profane' texts are marked with `1' or `0' harm potential, most of the `hate' texts were marked as `3' harm potential (or in some cases `2' as well). The offensive texts were marked with `2', `1' and even `0' harm potential. Table \ref{tab:comparison} illustrates the mapping between the two. In our comparison of the spans being marked using the HarmPot framework and those marked in the HateNorm task, we did not find many exact overlap of the spans selected in the two datasets. In most instances, the spans marked in the HateNorm task were a substring of those marked using the HarmPot framework and the total number of spans was also less in the HateNorm task - this is mainly because we do not mark hate spans, rather it's the identity spans that are marked in our framework. Level 2 of 2022 and 2023 editions, which mark whether it is contextual or standalone hate do not have a direct relationship to any of the levels in HarmPot - the main reason being that our definition of `context' is more rooted in how it is defined in discourse analysis and pragmatics as different socio-cultural factors affecting the interpretation of the text (and not just parent / previous text in the thread). 

\subsection{HarmPot and OLID}

The Offensive Language Identification Dataset (OLID) contains a collection of over 14k annotated English tweets using a three-level annotation framework. Level A distinguishes between Offensive and Non-offensive texts. At Level B offensive texts are further classified into targeted and untargeted insults. Level C categorises targeted insults into Individual, Group and Other targets \citep{zampieri-etal-2019-predicting}. 
It's a comparatively coarse-grained tagset but unlike the HASOC tagset, it addresses the two questions of `who' is being targeted and `magnitude' of the attack. 

For level A, the results of the comparative study were similar to the HASOC dataset. The rest of the two levels in the OLID framework relate to the `who' parameter in HarmPot but work at different axes - while OLID marks whether an individual or a group is being attacked, HarmPot looks at the specific identities irrespective of it being that of an individual or a group (more appropriately they represent the identity of an individual as a member of a group). All the texts with spans mentioning one of the identities and carrying a harm potential greater than `0' were marked as `Targeted Insult' at OLID's Level B. Any text without mention of any of the identities was mostly marked `Untargeted' - however, the magnitude of harm potential for such texts varied. This follows from the fact that the use of profane or unacceptable language may not necessarily trigger offline harm - one such instance could be friendly banter, which will have `0' harm potential. Texts marked as `Individual' or `Group' at Level C in the OLID dataset were marked for one of the identities in HarmPot. However, those marked as `Other' were not marked for identities (although the number of such texts was very small in the overall dataset). Also, the mapping of these categories to harm potential is quite unpredictable. The tentative mapping of OLID framework to HarmPot is summarised in Table \ref{tab:comparison}.

\subsection{HarmPot and ComMA}
Lastly we also conducted a comparative study between the HarmPot and ComMA framework \citep{kumar-etal-2018-aggression,kumar-EtAl:2022:LREC1}. 
The top level of the framework distinguishes between overtly, covertly and non-aggressive texts. At the second level, the aggression intensity of the aggressive texts 
- physical threat, sexual threat, non-threatening aggression and curse/abuse - 
are marked. Parallel to this, bias and threats of four kinds - religious, caste/class, gender and racial/ethnic - are marked. It also marks the discursive roles - attack, defend, counterspeech, abet and instigate and gaslighting - of the text. These discursive roles are already borrowed and incorporated in the HarmPot framework. Besides this, there are several parallels between the ComMA and HarmPot frameworks and also since social or physical `harm' is inherent to the idea of aggression, we expected a good mapping between the notion of verbal aggression and the harm potential of a text. 

The study showed that most of the non-aggressive texts (NAG) are at Level 0 but the vice-versa is not necessarily true. Also, most of the `covertly aggressive' (CAG) texts are categorised with level `1' harm potential. At the second level, physical and sexual threats were mostly marked as having `2' or `3' harm potential while non-threatening aggression is mostly marked as `1'. As in the earlier instances, some of the curse/abuse texts were also marked with `0' harm potential. At the level of threat and bias, even though religious, caste and gender bias have direct parallels in HarmPot since we are marking all mentions of these identities and not just biased or threatening ones (unlike the ComMA dataset), the instances of such spans were higher in our case. However, threats generally carried a harm potential of `2' or `3' while bias carried a harm potential of `1' or `2'. Some of the comments marked as non-biased in the ComMA dataset also carried a harm potential of `1' or even `2'. Moreover, the ComMA dataset marked the biases at the document level while HarmPot marks these at the span level.

\section{Conclusion}

In this paper, we have presented a new framework that could be used for annotating social media text with its potential for triggering offline harm. The framework incorporates contextual information such as the identity of the victim (as mentioned/referred to in the text), the broad socio-political situation in which the post is situated and the role that the text assumes in the discourse. We have also proposed using mood and modality as relevant categories for marking the speaker's intention, intended goal and their own evaluation of whether what they are saying is `necessary' or `possible'. These semantic categories have been rarely utilised in NLP but they could prove to be extremely useful in the identification of subjective phenomena like harm potential. We have annotated a total dataset of 4,000 texts - 2,000 related to the possible triggers of offline harm incidents and another 2,000 from datasets available for aggressive and hateful language identification. We use these 2,000 to carry out a comparative study of HarmPot with three popular frameworks and establish that a one-to-one mapping between these frameworks is not possible mainly because HarmPot does not mark hateful language; rather offline harm potential of the text. It shows some correlation between hateful language and its harm potential but neither entails the other. We are currently annotating some more data and also conducting experiments for the automatic identification of harm potential to understand the practical efficacy of the framework.

\section{Ethical Considerations}

The nature of the task - the creation of datasets with high harm potential and its annotation - in itself raises several ethical issues of bias and psychological impact on the annotators working with the data. In order to reduce the impact of working with such data, we took 3 steps - (a) a `maximum' limit of 200 texts per week was set for the annotators - the annotators were barred from going through more than this number of texts in a week; (b) we had made arrangements for psychological counselling of the annotators working on the data; (c) a compulsory weekly `venting out' meeting was organised to enable annotators to talk to each other and other members of the project that allowed them to talk about, discuss and (hopefully) figure out the ridiculousness of the data that they were going through. We made a very conscious decision not to use crowdsourcing or even third-party annotators for data annotation and collection so as to ensure that these mechanisms are put in place.

In order to minimise the bias in the annotations and also make different perspectives on the data public, we have decided to release the disaggregated dataset with the annotations of all the annotators (with their disagreements). We were very conscious not to push for an agreement where it was not possible. Moreover, our in-house annotators were from mutually distinct socio-political, religious, cultural, and educational backgrounds, providing an innate cancelling out of any one type of bias overpowering the data analysis and interpretation - we have tried to annotate the data in such a way as to reflect different perspectives on the data (and not propound a single, homogeneous view).

\section{Limitations}
One of the primary limitations of the framework and the dataset is the lack of multimodal information being included in it. A large number of hateful and abusive language used on social media, with a high potential for harm, is expected to be accompanied by visuals including images and video. We are working on expanding the dataset to include multimodal data and see how well the framework adapts to that and also what kind of modifications would be needed for handling those cases. The second limitation is the pipeline-based workflow that the framework enforces, which has a greater chance of error propagation - if, for example, the system makes an error in recognising mood and modality, that might ultimately lead to an error in the prediction of harm potential itself. This is a general limitation of the hierarchical frameworks.

\nocite{*}
\section{Bibliographical References}\label{sec:reference}
\bibliographystyle{lrec-coling2024-natbib}
\bibliography{lrec-coling2024-example}


\end{document}